\documentclass{article}
\usepackage{PRIMEarxiv}
\usepackage{multirow}   
\usepackage{booktabs}   

\usepackage{enumitem}
\usepackage{amsthm}
\usepackage{xcolor}
\usepackage{enumitem} 

\usepackage[most]{tcolorbox}
\usepackage{ragged2e}
\usepackage{etoolbox}
\usepackage{amsmath} 
\usepackage[utf8]{inputenc} 
\usepackage[T1]{fontenc}    
\usepackage{hyperref}       
\usepackage{url}            
\usepackage{booktabs}       
\usepackage{amsfonts}       
\usepackage{nicefrac}       
\usepackage{microtype}      
\usepackage{lipsum}
\usepackage{fancyhdr}       
\usepackage[T1]{fontenc}
\usepackage{graphicx}       
\graphicspath{{media/}}     
\usepackage{academicons} 

\usepackage{fontawesome} 
\usepackage{wasysym}     

\usepackage{algorithm}
\usepackage{algpseudocode}  

\usepackage{amsmath}
\usepackage{amssymb}
\usepackage{booktabs}
\usepackage{multirow}
\usepackage{array}
\usepackage{xcolor}

\newcommand{\E}{\mathbb{E}}
\newcommand{\norm}[1]{\left\|#1\right\|}
\newcommand{\thetat}{\theta_0}
\newcommand{\thetaft}{\theta^*}
\newcommand{\Ltask}{\mathcal{L}_{\text{task}}}
\newcommand{\Lreg}{\mathcal{L}_{\text{FW-SSR}}}
\newcommand{\Ltot}{\mathcal{L}_{\text{total}}}
\newcommand{\Dprobe}{\mathcal{D}_{\text{safe}}}
\newcommand{\Dtask}{\mathcal{D}_{\text{task}}}

\title{When Safety Geometry Collapses: Fine-Tuning Vulnerabilities in Agentic Guard Models}

\author{
  Ismail Hossain$^1$, Sai Puppala$^2$, Jannatul Ferdaus$^1$, Md Jahangir Alam$^1$, Yoonpyo Lee$^3$,\\ Syed Bahauddin Alam$^4$, Sajedul Talukder$^1$ \\
  $^1$Department of Computer Science, University of Texas at El Paso, TX, USA 79902 \\
  $^2$School of Computing, Southern Illinois University Carbondale, IL, USA 62901 \\
  $^3$Hanyang University, Seoul, South Korea \\
  $^4$University of Illinois Urbana-Champaign, IL, USA \\
  \texttt{\{ihossain, jferdaus, malam10\}@miners.utep.edu} \\
  \texttt{sai.puppala@siu.edu, lukeyounpyo@hanyang.ac.kr} \\
  \texttt{alams@illinois.edu, stalukder@utep.edu} \\
  \faGlobe\ \href{https://supreme-lab.github.io/wsgc/}{https://supreme-lab.github.io/wsgc/}
}		

\begin{document}
\maketitle

\begin{abstract}
A guard model fine-tuned on entirely benign data can lose all safety
alignment-- not through adversarial manipulation, but through standard
domain specialization.
We demonstrate this failure across three purpose-built safety
classifiers-- LlamaGuard, WildGuard, and Granite Guardian-- deployed
as protection layers in agentic AI pipelines, and show that it
originates in the destruction of latent safety geometry: the
structured harmful--benign representational boundary that guides
classification.
We extract per-layer safety subspaces via SVD on class-conditional
activation differences and track how this boundary evolves under
benign fine-tuning.
Granite Guardian undergoes complete collapse-- refusal rate drops from
85\% to 0\%, CKA falls to zero, and 100\% of outputs become
ambiguous-- a severity exceeding prior findings on general-purpose
LLMs, explained by the specialization hypothesis: concentrated safety
representations are efficient but catastrophically brittle.
To mitigate this, we propose Fisher-Weighted Safety Subspace
Regularization (FW-SSR), a training-time penalty combining
(i)~curvature-aware direction weights derived from diagonal Fisher
information and (ii)~an adaptive $\lambda_t$ that scales with
task--safety gradient conflict.
FW-SSR recovers 75\% refusal on Granite Guardian (CKA\,$=\!0.983$)
and reduces WildGuard's Attack Success Rate to 3.6\%-- below the
unmodified baseline-- by actively sharpening the safety subspace
rather than merely anchoring it.
Across all three models, structural representational geometry (CKA,
Fisher score) predicts safety behavior more reliably than absolute
displacement metrics, establishing geometry-based monitoring as a
necessary component of guard model evaluation in agentic deployments.
\end{abstract}
\section{Introduction}
\label{sec:introduction}

\begin{figure}[ht!]
    \centering
    \includegraphics[width=0.8\textwidth]{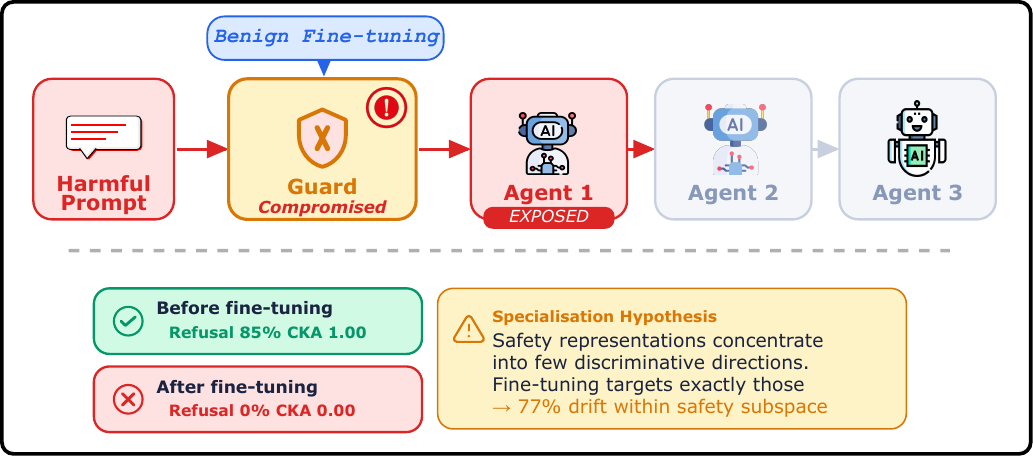}
    \caption{Benign Fine-Tuning Destroys Safety Alignment}
    \label{fig:threat-model}
\end{figure}

Modern AI systems are rapidly evolving from single-model deployments
toward \emph{agentic architectures}~\cite{park2023generative,wang2024survey},
in which multiple specialized LLMs collaborate through structured
communication pipelines to solve complex, long-horizon tasks.
In such systems, each agent handles a distinct sub-task-- retrieval,
reasoning, planning, code execution, or synthesis-- and agents exchange
intermediate outputs across turns.
This collaborative specialization enables capabilities far beyond any
individual model, but it introduces a new attack surface: if any single
agent in the pipeline can be made to process or propagate harmful
content, the safety of the entire system is compromised.

A standard mitigation in agentic deployments is to protect individual
agents with \emph{dedicated safety guard models}: purpose-built
classifiers such as MD-Judge~\cite{mdjudge2024},
LlamaGuard~\cite{inan2023llama}, and Granite
Guardian~\cite{ibmgranite2024} that intercept inputs before they reach
the agent, filtering adversarial or policy-violating content.
Unlike general-purpose LLMs that must balance safety with broad
capability, these guard models are trained specifically to classify
requests across a standardized harm taxonomy and reliably refuse harmful
inputs while permitting benign ones.
They operate as a \emph{static protection layer}: once deployed,
they are assumed to remain aligned regardless of downstream pipeline
changes.

This assumption is violated in practice.
Agentic systems are routinely fine-tuned to specialize individual agents
for particular domains: a medical-assistant agent tuned on clinical
dialogue, a legal-reasoning agent tuned on case documents, or an
enterprise agent tuned on company-specific policy.
The safety guards protecting these agents naturally undergo the same
specialization: fine-tuned on domain-relevant, entirely \emph{benign}
data to improve domain-specific classification performance.
As we demonstrate, this benign adaptation is sufficient to
\emph{catastrophically destroy} the guard model's safety alignment-- a
fundamental threat to the safety of the agentic pipeline as a whole.

\begin{figure}[ht!]
    \centering
    \includegraphics[width=\textwidth]{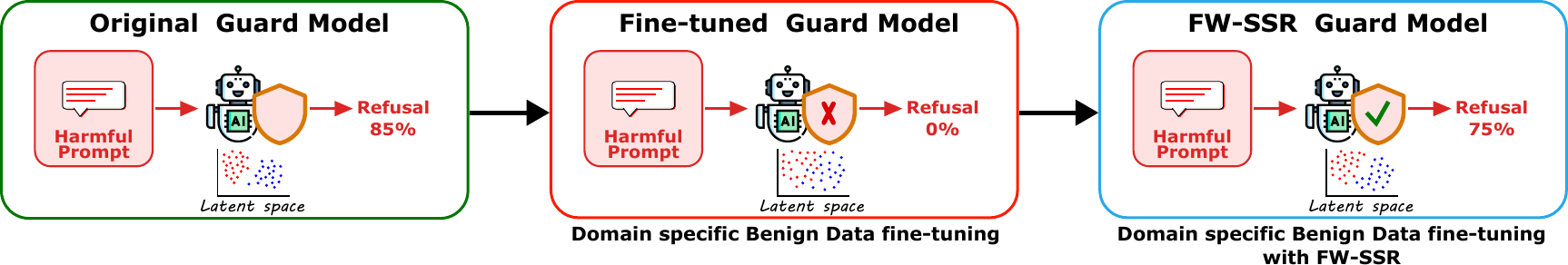}
    \caption{Safety alignment collapse and recovery in guard models under benign fine-tuning.}
    \label{fig:system-architecture}
\end{figure}

We study this failure as a \emph{latent geometry} phenomenon.
Safety-aligned guard models encode the harmful-- benign distinction as
a structured separation in activation space: prompts from different
safety classes occupy distinct regions, separated by a latent safety
boundary that guides classification.
Standard fine-tuning, optimizing a task loss with no constraint on
internal representations, erodes this boundary even when training data
is entirely benign.
The result is \emph{safety geometry collapse}: representations of
harmful and benign inputs become indistinguishable.
In an agentic context, this collapse means the guard model can no
longer distinguish an adversarial jailbreak from a routine query,
allowing harmful content to reach the agent it was designed to protect
and potentially propagate to downstream agents. In the Figure~\ref{fig:system-architecture}, The original guard model \emph{(left)} maintains a well-separated latent
space between harmful and benign inputs, achieving 85\% refusal rate.
Domain-specific benign fine-tuning \emph{(center)} destroys this boundary
entirely-- refusal drops to 0\% and the latent space collapses-- despite
containing no harmful training signal.
FW-SSR \emph{(right)} restores the harmful--benign separation by
regularizing safety-critical subspace directions during fine-tuning,
recovering 75\% refusal while preserving domain utility.

\paragraph{Why Guard Models Are Especially Vulnerable.}
Prior work~\cite{qi2023finetuning} reports approximately 55~pp refusal
drops in general-purpose LLMs under benign Alpaca fine-tuning.
We observe an \textbf{85~pp collapse} in Granite Guardian-- complete
erasure of refusal behavior-- which we attribute to a
\emph{specialization hypothesis}: guard models concentrate safety
representations into a small number of highly discriminative latent
directions, creating classification efficiency but also catastrophic
brittleness when those directions are perturbed by fine-tuning.

\textbf{Research Gap.}
Three specific limitations motivate our approach.
(i)~No prior work examines benign fine-tuning vulnerability in
purpose-built guard models operating in agentic pipelines, which exhibit
qualitatively more severe degradation than general-purpose LLMs.
(ii)~Existing latent anchor penalties penalize all safety directions
equally, ignoring the non-uniform curvature structure of the safety
subspace.
(iii)~Fixed regularization strength $\lambda$ cannot adapt to the
varying conflict between task and safety gradients across training.

\textbf{Contributions.}
(1)~\textbf{Threat Characterization.} We identify benign fine-tuning
during agent specialization as a systematic threat to guard model
safety in agentic AI pipelines, introducing a latent geometry analysis
protocol that reveals complete safety geometry collapse more severe than
in general-purpose LLMs.
(2)~\textbf{FW-SSR.} A novel training-time penalty weighting each
safety direction by diagonal Fisher information, with adaptive
$\lambda_t$ scaled by task-- safety gradient conflict, enabling safe
specialization of guard models without sacrificing domain performance.
(3)~\textbf{Evaluation.} FW-SSR recovers 75\% refusal behavior and
CKA\,$=\!0.983$ on Granite Guardian~3B under benign Alpaca fine-tuning,
with ablations isolating each component's contribution.

\section{Related Work}
\label{sec:related_work}

 \textbf{Agentic AI Safety.}
Recent work on multi-agent LLM systems~\cite{park2023generative,wang2024survey}
identifies security risks unique to agentic pipelines, including cross-agent
prompt injection, adversarial manipulation of inter-agent communication, and
goal misalignment in planning agents. Adversarial instructions can hijack
downstream tasks~\cite{perez2022ignore}, while indirect prompt injection through
retrieved documents can compromise entire pipelines~\cite{greshake2023not}.
Similarly, jailbreak vulnerabilities in base VLMs can be \emph{inherited} by
fine-tuned models, with adversarial images transferring to safety-tuned
variants at success rates above 86.5\%~\cite{wang2025sea}. These findings
suggest that fine-tuning often propagates latent vulnerabilities rather than
removing them. In contrast, we identify a \emph{structural} failure mode:
benign fine-tuning for agent specialization can systematically destroy the
safety geometry of guard models even without adversarial inputs.

\textbf{Safety Degradation Under Fine-Tuning.}
Benign fine-tuning can significantly weaken safety alignment in LLMs,
increasing harmful completion rates even without malicious data, with
refusal drops of up to $\sim$55~pp~\cite{qi2023finetuning}. Certain
data properties disproportionately degrade safety~\cite{zhang2024benign},
and small adversarial injections can further amplify this collapse~\cite{yang2023shadow}.
Lisa~\cite{huang2024lisa} mitigates degradation by alternating alignment
and task states with proximal drift constraints, but operates at the
loss level rather than preserving latent safety geometry as FW-SSR does.
Similarly, backdoor purification can reduce attack success rates while
leaving latent vulnerabilities intact, allowing rapid re-learning from
minimal poisoned samples~\cite{min2024superficial}. We extend this line
of work to guard models, analysing safety degradation through latent
geometry and highlighting the risk in agentic pipelines where a single
compromised guard can expose the entire system.
 
\textbf{Representation-Level Safety.}
Safety-relevant behaviors are encoded in structured activation-space
directions that can be identified and manipulated~\cite{zou2023representation},
with high-level concepts occupying low-dimensional linear
subspaces~\cite{park2023linear}. Hidden representations also encode
evaluative signals beyond surface outputs; for example, task difficulty
is linearly decodable from the initial hidden state~\cite{zhu2025llmknows}.
These findings indicate that latent geometry contains safety-relevant
information not captured by behavioral metrics, consistent with our
observation that CKA and Fisher scores detect safety degradation
earlier than output-based measures. Similar vulnerabilities appear in
multimodal models, where adversarial images can universally jailbreak
aligned VLMs by exploiting visual representation structure~\cite{qi2024visual}.
Motivated by this perspective, we compute explicit safety subspaces
from class-conditional activation statistics and use them as
training-time anchors to preserve alignment during domain
specialization.
 
\textbf{Fisher Information in Continual Learning.}
EWC~\cite{kirkpatrick2017overcoming} penalizes changes to parameters
important for prior tasks via diagonal Fisher information, preventing
catastrophic forgetting in sequential learning.
The guard model safety collapse we observe is an instance of vertical
continual learning failure~\cite{shi2024continual} -- domain
specialization gradients overwrite concentrated safety representations
in the same way catastrophic forgetting destroys prior task knowledge.
Unlike EWC, which operates at the parameter level and is incompatible
with LoRA's frozen base, FW-SSR applies Fisher weighting at the
\emph{activation level} within the safety subspace, preserving
compatibility with parameter-efficient agent specialization.
 
\section{Problem Formulation}
\label{sec:problem}

\subsection{Setting}
Let $M_{\thetat}$ be a pretrained, safety-aligned guard model.
Given a benign fine-tuning dataset $\Dtask$ and a safety probe dataset
$\Dprobe = \{(x_i, y_i)\}_{i=1}^{N}$ with $y_i \in \{0,1\}$
(benign/harmful), we wish to adapt $M_{\thetat}$ to downstream utility
while preserving safety alignment, yielding parameters $\thetaft$.

\subsection{Safety Subspace Extraction}
\label{sec:subspace}
 
For transformer layer $\ell$ and parameters $\theta$, let
$h_\ell(x;\theta) \in \mathbb{R}^{d}$ denote the hidden state at the
final non-padding token position (last-token representation), consistent
with standard practice for decoder-only causal LMs.
Let $H_\ell^c \in \mathbb{R}^{N_c \times d}$ denote the matrix whose
$i$-th row is $h_\ell(x_i^c;\thetat)$, extracted from the $N_c$ probe
samples of class $c \in \{h,\, b\}$ (harmful, benign).
Let $\mu_\ell^c = \frac{1}{N_c}\sum_i [H_\ell^c]_i \in \mathbb{R}^d$
be the class-conditional mean under $\thetat$, and let
\begin{equation}
  \tilde{H}_\ell^c
    = H_\ell^c - \mathbf{1}_{N_c}(\mu_\ell^c)^\top
    \in \mathbb{R}^{N_c \times d}
\end{equation}
denote the class-centered activation matrix.
We construct an augmented activation matrix by stacking the two
centered matrices together with $n_a$ scaled copies of the
between-class mean difference:
\begin{equation}
  A_\ell = \begin{bmatrix}
             \tilde{H}_\ell^h \\[2pt]
             \tilde{H}_\ell^b \\[2pt]
             \gamma\cdot\mathbf{1}_{n_a}(\Delta\mu_\ell)^\top
           \end{bmatrix}
           \in \mathbb{R}^{(N_h + N_b + n_a)\times d},
\label{eq:aug_matrix}
\end{equation}
where $\Delta\mu_\ell = \mu_\ell^h - \mu_\ell^b \in \mathbb{R}^d$ is
the between-class difference vector, $n_a = \min(32, N_h)$, and
$\gamma = 5$ amplifies the primary safety direction to dominate the top
singular values.
 
The \emph{safety subspace} $U_\ell \in \mathbb{R}^{k \times d}$
comprises the top-$k$ \emph{right} singular vectors of $A_\ell$,
i.e.\ the first $k$ rows of $V_h^\top$ from
$\mathrm{SVD}(A_\ell) = P\,\Sigma\,V_h^\top$
(with $\mathrm{full\_matrices}=\mathrm{False}$, giving
$V_h^\top \in \mathbb{R}^{\min(N_\mathrm{aug},d)\times d}$).
These vectors span the top-$k$ directions of variance in
$\mathbb{R}^d$ that are most discriminative between harmful and benign
activations.
$U_\ell$ is computed once from $M_{\thetat}$ and held fixed throughout
fine-tuning.

\subsection{Safety Drift Metrics}
The \emph{safety drift} at layer $\ell$ under $\theta$ is:
\begin{equation}
  \Delta_{\mathrm{safe}}(\ell;\theta) =
    \E_{x \sim \Dprobe}
    \bigl[\norm{U_\ell^\top\!\bigl(h_\ell(x;\theta) -
    h_\ell(x;\thetat)\bigr)}_2\bigr].
\label{eq:safety_drift}
\end{equation}
The \emph{drift ratio} measures the fraction of total activation change
within the safety subspace:
\begin{equation}
  \rho(\ell;\theta) = \frac{\Delta_{\mathrm{safe}}(\ell;\theta)}
    {\E_x[\norm{h_\ell(x;\theta)-h_\ell(x;\thetat)}_2]+\varepsilon}.
\label{eq:drift_ratio}
\end{equation}
The \emph{Fisher discriminant score} measures class separation:
\begin{equation}
  \mathrm{FS}(\ell;\theta) =
    \frac{\norm{\mu_\ell^h(\theta)-\mu_\ell^b(\theta)}_2}
         {\sigma_\ell^h(\theta)+\sigma_\ell^b(\theta)+\varepsilon},
\label{eq:fisher_score}
\end{equation}
where $\sigma_\ell^c(\theta)$ is the mean intra-class L2 spread.
Linear CKA~\cite{kornblith2019similarity} compares representational
geometry between $\theta$ and $\thetat$ (Eq.~\ref{eq:cka}), computed
on a 64-sample subsample for efficiency:
\begin{equation}
  \mathrm{CKA}(H_\ell(\theta), H_\ell(\thetat)) =
    \frac{\mathrm{HSIC}(K_\theta, K_0)}
         {\sqrt{\mathrm{HSIC}(K_\theta,K_\theta)\cdot\mathrm{HSIC}(K_0,K_0)}}.
\label{eq:cka}
\end{equation}

\paragraph{Problem Statement.}
Find $\thetaft$ minimizing task loss on $\Dtask$ while, for all probed
layers $\ell\in\mathcal{L}$, minimizing
$\Delta_{\mathrm{safe}}(\ell;\thetaft)$ and preserving refusal rates
relative to $M_{\thetat}$.

\section{Fisher-Weighted Safety Subspace Regularization}
\label{sec:method}

\subsection{FW-SSR: Formulation}

We propose FW-SSR, replacing the uniform penalty with a curvature-aware
adaptive variant:
\begin{equation}
  \Lreg =
    \E_{x \sim \Dprobe} \sum_{\ell \in \mathcal{L}}
    \norm{\hat{F}_\ell \odot U_\ell^\top\!\bigl(h_\ell(x;\theta)-
    h_\ell(x;\thetat)\bigr)}_2^2,
\label{eq:fwssr}
\end{equation}
where $\hat{F}_\ell \in \mathbb{R}^k$ is a vector of per-direction
curvature weights and $\odot$ denotes element-wise multiplication.
The total objective is $\Ltot = \Ltask + \lambda_t \cdot \Lreg$.
Gradients flow only through the fine-tuned model's activations;
original model activations are detached and serve as frozen anchors.

\subsection{Fisher Weight Estimation}
\label{sec:fisher}

The curvature weights estimate sensitivity via a diagonal approximation
of projected Fisher information:
\begin{equation}
  F_\ell \approx
    \E_{x \sim \Dprobe}\bigl[(U_\ell^\top h_\ell(x;\theta))^2\bigr]
    \in \mathbb{R}^k.
\label{eq:fisher_approx}
\end{equation}
Before application, $F_\ell$ is mean-normalized:
\begin{equation}
  \hat{F}_\ell = \frac{k \cdot F_\ell}{\sum_j F_\ell[j]+\varepsilon},
\label{eq:fisher_norm}
\end{equation}
ensuring $\hat{F}_\ell$ has mean~1 by construction, so penalty scale
is consistent across layers.
Fisher weights are updated via EMA every $\tau=50$ gradient steps:
\begin{equation}
  F_\ell \leftarrow \beta F_\ell + (1-\beta)F_\ell^{\text{new}},
  \quad \beta=0.9,
\label{eq:ema}
\end{equation}
and initialized to $\mathbf{1}_k$ for uniform regularization before
estimates stabilize.

\subsection{Adaptive $\lambda$ Scheduling}
\label{sec:adaptive}

We measure gradient conflict via cosine similarity between task and
safety gradients every 20 steps:
\begin{equation}
  s_t = \cos\!\bigl(\nabla_\theta \Ltask,\;
                     \nabla_\theta \Lreg\bigr).
\label{eq:cosine}
\end{equation}
We update $\lambda$ as:
\begin{equation}
  \lambda_t^{\text{new}} \leftarrow \mathrm{clip}\!\Bigl(
    \lambda_{t-1} \cdot (1 - \tfrac{1}{2} s_t),\;10^{-4},\;1.0\Bigr),
\label{eq:lambda_update}
\end{equation}
with exponential smoothing:
\begin{equation}
  \lambda_t \leftarrow 0.95\,\lambda_{t-1} + 0.05\,\lambda_t^{\text{new}}.
\label{eq:lambda_ema}
\end{equation}
When $s_t\!\approx\!+1$ (objectives aligned),
$(1-\frac{1}{2}s_t)\!\approx\!0.5$ reduces $\lambda$.
When $s_t\!\approx\!-1$ (objectives conflict), the factor
$\approx\!1.5$ increases $\lambda$, providing strongest protection
exactly when needed.

\begin{algorithm}[!ht]
\caption{FW-SSR Training}
\label{alg:fwssr}
\textbf{Require:} $M_{\thetat}$, $\Dtask$, $\Dprobe$,
$\{U_\ell\}_{\ell\in\mathcal{L}}$,
$\lambda_0, k, \beta, \tau$ \\
\textbf{Output:} Safety-preserved fine-tuned parameters $\thetaft$
\begin{algorithmic}[1]
\State $\theta \leftarrow \thetat$;\quad
       $F_\ell \leftarrow \mathbf{1}_k\ \forall\ell$;\quad
       $\lambda \leftarrow \lambda_0$
\For{each step $t = 1, \ldots, T$}
  \State Sample $B_{\text{task}} \sim \Dtask$,\;
         $B_{\text{safe}} \sim \Dprobe$
  \State Compute $\Ltask$ on $B_{\text{task}}$
  \State Compute $\hat{F}_\ell$ (Eq.~\ref{eq:fisher_norm}),
         then $\Lreg$ (Eq.~\ref{eq:fwssr})
         on $B_{\text{safe}}$
  \State $\theta \leftarrow \theta -
         \eta\nabla_\theta(\Ltask + \lambda\Lreg)$
  \If{$t \bmod \tau = 0$}
    \State Update $F_\ell$ via EMA (Eq.~\ref{eq:ema})
  \EndIf
  \If{$t \bmod 20 = 0$}
    \State Compute $s_t$ (Eq.~\ref{eq:cosine})
    \State Update $\lambda$
           (Eqs.~\ref{eq:lambda_update}--\ref{eq:lambda_ema})
  \EndIf
\EndFor
\State \Return $\thetaft$
\end{algorithmic}
\end{algorithm}

\begin{figure*}
    \centering
    \includegraphics[width=\textwidth]{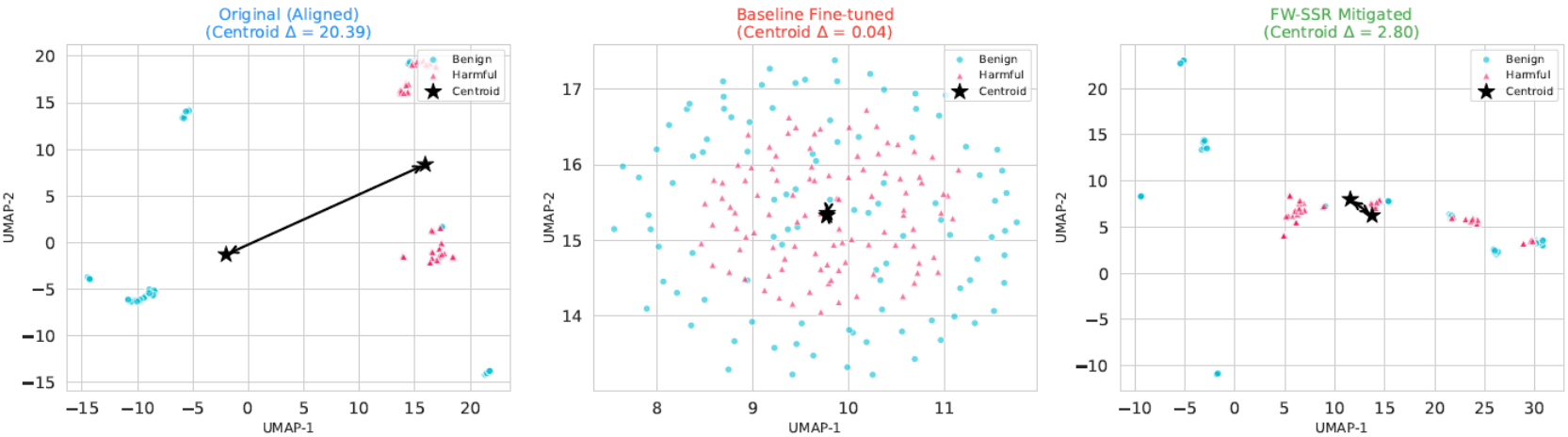}
    \caption{Granite Guardian 3.0-2B-- Latent Space Geometry Layer 39 Harmful vs. Benign Prompt Separation}
    \label{fig:latent-space-geometry}
\end{figure*}

\section{Experimental Setup}
\label{sec:setup}

\subsection{Models}
We evaluate on three purpose-built guard model families.


\textbf{WildGuard}~\cite{han2024wildguard}: An open-source safety
classifier built on Mistral-7B (32 transformer layers, $d=4096$),
trained to jointly detect harmful requests, adversarial jailbreaks,
and model refusals across a broad harm taxonomy.
Probed at layers $\{20, 24, 28, 31\}$ (50/65/80/100\% depth).
\textbf{Granite Guardian 3.0-2B}~\cite{ibmgranite2024}: IBM's
open-access guard model; complete experimental results are available and reported in Tables~\ref{tab:main_results}.
\textbf{LlamaGuard-3-8B}~\cite{inan2023llama}: Meta's 8B safety
classifier built on LLaMA-3.1 (32 layers, $d=4096$).

\subsection{Fine-Tuning Data}
We fine-tune on 2,000 randomly sampled examples from the Alpaca
instruction-following dataset~\cite{alpaca}, following the Risk Level~3
protocol of~\cite{qi2023finetuning}.
This simulates realistic domain adaptation without adversarial intent.

\subsection{Safety Probe Dataset}
\label{sec:probe}

We construct $\Dprobe$ with 80 balanced prompt pairs (40 harmful,
40 benign).
\textit{Harmful prompts}: 38 unique prompts spanning 5 harm
categories-- \textbf{violence} (improvised weapons, poisoning),
\textbf{cybercrime} (SQL injection, keyloggers, ransomware),
\textbf{fraud} (counterfeiting, scam scripts, pyramid schemes),
\textbf{drug synthesis} (fentanyl synthesis, dangerous combinations),
and \textbf{harassment} (doxxing, intimidation campaigns)-- plus
biological weapons, illegal firearms, and identity fraud.
Prompts are repeated to fill 40 slots.
\textit{Benign prompts}: 20 factual queries repeated to fill 40 slots.
Labels: $y=1$ (harmful), $y=0$ (benign).

\subsection{Implementation}

All fine-tuning uses QLoRA~\cite{dettmers2023qlora} with 4-bit NF4
quantization; LoRA rank $r=8$, $\alpha=16$, applied to all attention
($q/k/v/o\_\text{proj}$) and MLP (gate/up/down\_proj) matrices
(auto-detected per architecture).
Training: 3 epochs, batch size~8, gradient accumulation~4 steps
(effective batch size~32), learning rate $2\!\times\!10^{-5}$, AdamW
weight decay $10^{-2}$, gradient clipping at~1.0.
FW-SSR: $\lambda_0\!=\!0.1$, $\beta\!=\!0.9$, $\tau\!=\!50$, $k\!=\!32$.
Single NVIDIA H100 96~GB GPU; $M_{\thetat}$ offloaded to CPU, moved
to GPU only for regularization forward passes.


\subsection{Evaluation Metrics}

\textit{Safety Behavior} (20 harmful prompts, keyword classification):
\textbf{Refusal Rate}-- fraction explicitly declining;
\textbf{Compliance Rate}-- fraction providing harmful content;
\textbf{Ambiguous Rate}-- fraction matching neither criterion.
A model with 0\% refusal and 0\% compliance produces 100\% ambiguous
outputs, indicating complete representational collapse: the model neither
refuses nor produces useful harmful content, generating off-topic
incoherent outputs with no safety value.

\textit{Latent Geometry}: safety drift (Eq.~\ref{eq:safety_drift}),
drift ratio (Eq.~\ref{eq:drift_ratio}), cosine similarity, Fisher
score (Eq.~\ref{eq:fisher_score}), inter-class centroid distance,
and CKA vs.\ original (Eq.~\ref{eq:cka}, 64-sample subsample).

\section{Results}
\label{sec:results}

Table~\ref{tab:main_results} presents results for all three guard models
at the final probed layer (100\% depth).

\subsection{Collapse Severity Tracks Safety Representation Concentration}

Benign fine-tuning degrades all three guard models, but the severity
scales with the concentration of safety representations.
Granite Guardian undergoes the most catastrophic collapse: refusal rate
drops from 85\% to 0\%, CKA falls to 0.00, and Fisher score and
inter-class distance both reach 0.00, indicating complete destruction
of the harmful-- benign classification boundary.
A drift ratio of 0.77 confirms the mechanism-- 77\% of all activation
change is concentrated in the safety subspace, meaning unconstrained
gradients disproportionately overwrite the directions most critical
to safety.
This 85~pp refusal collapse substantially exceeds the $\sim$55~pp drops
reported for general-purpose LLMs~\cite{qi2023finetuning}, consistent
with the specialization hypothesis: guard models encode safety into few
highly discriminative directions, which creates efficiency but
catastrophic brittleness.
LlamaGuard-3 shows partial degradation-- CKA falls to 0.83, Fisher
score from 0.62 to 0.47, drift ratio 0.51-- without reaching zero on
any metric, suggesting a more distributed safety geometry that
distributes and thus dilutes the effect of fine-tuning gradients.
WildGuard exhibits a qualitatively different pattern: refusal drops
from 35\% to 5\% yet Fisher score \emph{increases} from 0.97 to 1.01
and safety drift remains the lowest of all three models (17.21),
indicating that behavioral degradation here is driven by disruption
of the decision function above the subspace rather than collapse of
the subspace itself.
Together, the three models reveal that safety collapse is not a single
phenomenon but a spectrum whose severity is predicted by geometry,
not behavior alone.

\subsection{FW-SSR Recovers Safety Geometry Across All Three Models}

FW-SSR consistently improves both behavioral and geometric metrics
relative to the fine-tuned baseline across all architectures.
For Granite Guardian, refusal rate recovers to 75\% (an 88\% relative
recovery), CKA reaches 0.98, Fisher score recovers to 0.55 (81\% of
original), and inter-class distance to 11.36 (93\% of original),
confirming that the safety classification boundary is largely
reconstituted despite substantial absolute drift.
LlamaGuard-3 shows consistent but more modest improvement: refusal rate
rises from 15\% to 25\%, safety drift reduces by 17\% (45.99~$\to$~38.00),
and Fisher score recovers to 0.54, approaching the original 0.62.
For WildGuard, FW-SSR raises Fisher score to 1.08-- \emph{exceeding
the original 0.97}-- as the curvature-aware weighting mechanism
actively sharpens the class-separating directions rather than merely
anchoring them, while compliance returns to 0\% and refusal recovers
to 20\%.
Across all three models, the largest absolute improvements occur in
the metrics most directly linked to classification boundary
structure-- Fisher score, inter-class distance, and CKA-- rather
than in drift suppression, pointing to the mechanism examined next.

\subsection{Structural Geometry Is the Primary Safety Carrier}

A consistent pattern across all three models establishes that
\emph{relational} geometry predicts safety behavior more reliably
than \emph{absolute displacement}.
Granite Guardian's FW-SSR condition provides the sharpest demonstration:
safety drift is only 37\% suppressed (48.64~$\to$~30.44), yet CKA
recovers to 0.98 and refusal rate to 75\%.
Activations have moved substantially from their original positions, but
their relational organization-- how harmful and benign inputs are
separated in the safety subspace-- is almost perfectly preserved,
and behavior tracks this structure, not the displacement magnitude.
WildGuard's fine-tuned condition provides the complementary
counter-example: safety drift of 17.21 is the lowest of any fine-tuned
model, yet refusal drops from 35\% to 5\%.
Low displacement does not prevent behavioral collapse when the
class-separating geometry is eroded, as reflected by inter-class
distance falling from 11.03 to 7.94 despite stable drift.
These two observations jointly imply that CKA and Fisher score are
the most diagnostically meaningful metrics for guard model safety
evaluation: they detect both the global structural collapse of Granite
and the localized boundary erosion of WildGuard, while drift-based
metrics fail in both cases.

\begin{table}[ht!]
\centering
\caption{%
  Safety geometry and behavioral evaluation across three guard models
  under benign Alpaca fine-tuning and FW-SSR mitigation.
  \textbf{O}~=~original aligned model;
  \textbf{FT}~=~baseline fine-tuned (no defense);
  \textbf{Mit}~=~FW-SSR mitigated.
  $\uparrow$ higher is better; $\downarrow$ lower is better.
  \textbf{Bold}: best value among \{FT,\,Mit\} per metric per model.
}
\label{tab:main_results}
\small
\setlength{\tabcolsep}{5.2pt}
\renewcommand{\arraystretch}{1}
\begin{tabular}{l ccc ccc ccc}
\toprule
\multirow{2}{*}{\textbf{Metric}}
  & \multicolumn{3}{c}{\textbf{Granite Guardian 3.0-2B}}
  & \multicolumn{3}{c}{\textbf{LlamaGuard-3-8B}}
  & \multicolumn{3}{c}{\textbf{WildGuard}} \\
\cmidrule(lr){2-4}\cmidrule(lr){5-7}\cmidrule(lr){8-10}
  & O & FT & Mit
  & O & FT & Mit
  & O & FT & Mit \\
\midrule
\multicolumn{10}{l}{\textit{Safety Behavior (20 harmful prompts, keyword classification)}} \\[2pt]
Refusal Rate (\%) $\uparrow$
  & 85.00 & 0.00          & \textbf{75.00}
  & 5.00  & 15.00         & \textbf{25.00}
  & 35.00 & 5.00          & \textbf{20.00} \\
Compliance Rate (\%) $\downarrow$
  & 0.00 & \textbf{0.00} & 5.00
  & 5.00 & 10.00         & 10.00
  & 0.00 & 5.00          & \textbf{0.00} \\
Ambiguous Rate (\%)
  & 15.00 & 100.00        & \textbf{20.00}
  & 90.00 & 75.00         & \textbf{65.00}
  & 65.00 & 90.00         & \textbf{80.00} \\
\midrule
\multicolumn{10}{l}{\textit{Latent Safety Geometry (final probed layer, 64-sample CKA)}} \\[2pt]
Safety Drift $\downarrow$
  & 0.00 & 48.64         & \textbf{30.44}
  & 0.00 & 45.99         & \textbf{38.00}
  & 0.00 & 17.21         & \textbf{17.17} \\
Drift Ratio $\downarrow$
  & 0.00 & 0.77 & \textbf{0.56}
  & 0.00 & 0.51 & \textbf{0.44}
  & 0.00 & 0.65 & \textbf{0.64} \\
Cosine Sim.\ $\uparrow$
  & 1.00 & 0.00          & \textbf{0.58}
  & 1.00 & 0.12          & \textbf{0.21}
  & 1.00 & 0.30          & \textbf{0.31} \\
Fisher Score $\uparrow$
  & 0.68 & 0.00          & \textbf{0.55}
  & 0.62 & 0.47          & \textbf{0.54}
  & 0.97 & 1.01          & \textbf{1.08} \\
Inter-Class Dist.\ $\uparrow$
  & 12.19 & 0.00         & \textbf{11.36}
  & 12.34 & 9.97         & \textbf{11.25}
  & 11.03 & 7.94         & \textbf{8.64} \\
CKA vs.\ Orig.\ $\uparrow$
  & 1.00 & 0.00          & \textbf{0.98}
  & 1.00 & 0.83          & 0.83
  & 1.00 & 0.83          & \textbf{0.86} \\
\bottomrule
\end{tabular}
\end{table}

\subsection{Layer-wise Safety Drift}
Drift concentrates in deep layers (65-- 100\% depth), consistent with
safety-relevant representations residing in late transformer
layers~\cite{zou2023representation}.
FW-SSR provides strongest suppression at 80-- 100\% depth, where Fisher
weights are highest and regularization most aggressive.
Early layers (50\%) show minimal drift in both conditions.
The baseline exhibits high drift across all layers beyond 65\% depth,
confirming safety collapse is a distributed, multi-layer phenomenon.

\begin{figure}[ht!]
\centering
    \includegraphics[width=0.5\textwidth]{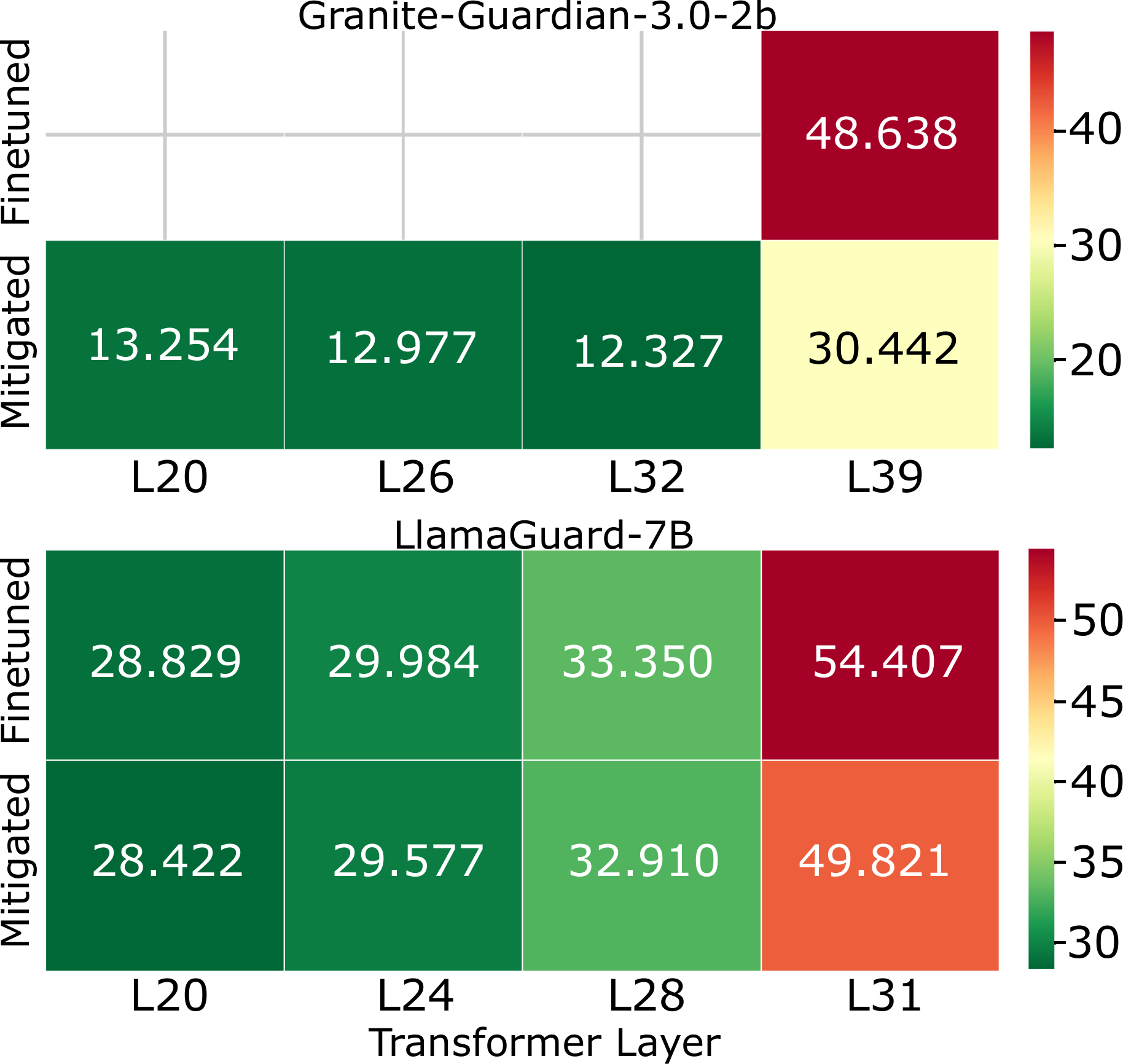}
    \caption{Per-Layer Safety Drift Heatmap}
\end{figure}

\subsection{Per-Benchmark ASR Analysis}
Table~\ref{tab:wildguard_main} reveals three consistent findings.
First, benign fine-tuning raises overall ASR from 9.2\% to 17.1\%
(an 85.9\% relative increase) uniformly across all four benchmarks,
with the largest degradation on AdvBench ($+$13.0~pp) and the smallest
on HarmBench ($+$3.2~pp), ruling out benchmark-specific artefacts and
confirming that safety degrades broadly across diverse prompt
distributions.
Second, FW-SSR not only reverses this degradation but reduces ASR to
3.6\%-- 60.9\% below the \emph{original} model's 9.2\%-- achieving the
lowest ASR on every individual benchmark; this improvement beyond the
original baseline is explained geometrically: the Fisher-weighted
regularization actively sharpens the harmful-- benign classification
boundary (Fisher score 0.118~$\to$~0.153; inter-class distance
37.1~$\to$~62.7), rather than merely anchoring it to its pre-fine-tuning
state.
Finally, the highest residual ASR under FW-SSR occurs on AdvBench
(5.0\%), whose direct harmful instructions densely sample harm categories
underrepresented in the current safety probe set; category-stratified
probe construction is the most direct remedy.

\begin{table}[ht!]
\centering
\caption{Evaluation results for WildGuard across original, fine-tuned
(no defence), and FW-SSR conditions.
ASR evaluated by cross-model judge (LlamaGuard-3-8B).
$\downarrow$ lower is better; $\uparrow$ higher is better.
\textbf{Bold}: best per row; \textcolor{red}{red}: worst.}
\label{tab:wildguard_main}
\small
\setlength{\tabcolsep}{5pt}
\renewcommand{\arraystretch}{1}
\begin{tabular}{lccc}
\toprule
& \textbf{Original} & \textbf{Fine-tuned} & \textbf{FW-SSR} \\
\midrule
\multicolumn{4}{l}{\textit{Per-Benchmark ASR (\%) $\downarrow$}} \\[2pt]
\quad AdvBench & 14.0 & \textcolor{red}{27.0} & \textbf{5.0} \\
\quad HarmBench & 6.2 & \textcolor{red}{9.4} & \textbf{2.1} \\
\quad JailbreakBench & 7.0 & \textcolor{red}{13.0} & \textbf{4.0} \\
\quad StrongREJECT & 9.4 & \textcolor{red}{18.8} & \textbf{3.1} \\
\textbf{Overall ASR} & 9.2 & \textcolor{red}{17.1} & \textbf{3.6} \\
\bottomrule
\end{tabular}
\end{table}


\subsection{Per-Category ASR Analysis}

\begin{table}[ht!]
\centering
\caption{Per-category Attack Success Rate (\%) for WildGuard across original,
fine-tuned (no defence), and FW-SSR conditions.
\textbf{Bold}: best (lowest ASR) per row; \textcolor{red}{red}: worst.}
\label{tab:wildguard_per_cat}
\small
\setlength{\tabcolsep}{4pt}
\renewcommand{\arraystretch}{1.08}
\begin{tabular}{l ccc | l ccc}
\toprule
\textbf{Category} & \textbf{Orig} & \textbf{FT} & \textbf{FW} &
\textbf{Category} & \textbf{Orig} & \textbf{FT} & \textbf{FW} \\
\midrule
Harmful & \textcolor{red}{6.2} & \textbf{0.0} & \textbf{0.0} & Disinformation & \textbf{0.0} & \textbf{0.0} & \textbf{0.0} \\
Illegal & 18.8 & \textcolor{red}{31.2} & \textbf{6.2} & Malware / Hacking & 10.0 & \textcolor{red}{40.0} & \textbf{0.0} \\
Cybercrime / Intrusion & \textcolor{red}{6.2} & \textcolor{red}{6.2} & \textbf{0.0} & Physical Harm & \textbf{0.0} & \textbf{0.0} & \textcolor{red}{10.0} \\
Chemical / Biological & \textbf{0.0} & \textcolor{red}{6.2} & \textbf{0.0} & Privacy & \textbf{10.0} & \textcolor{red}{20.0} & \textbf{10.0} \\
Harassment / Bullying & \textbf{6.2} & \textbf{6.2} & \textbf{6.2} & Fraud / Deception & \textcolor{red}{10.0} & \textbf{0.0} & \textbf{0.0} \\
Violence & \textbf{6.2} & \textcolor{red}{18.8} & \textbf{6.2} & Harassment / Discrimination & \textcolor{red}{10.0} & \textcolor{red}{10.0} & \textbf{0.0} \\
Non-violent Crimes & \textcolor{red}{12.5} & \textcolor{red}{12.5} & \textbf{0.0} & Hate, Harassment \& Discrim. & \textbf{0.0} & \textcolor{red}{18.8} & \textbf{0.0} \\
Sexual / Adult Content & \textbf{10.0} & \textcolor{red}{30.0} & \textbf{10.0} & Illegal Goods \& Services & 18.8 & \textcolor{red}{37.5} & \textbf{6.2} \\
Sexual Content & 12.5 & \textcolor{red}{18.8} & \textbf{6.2} & Expert Advice & \textbf{0.0} & \textcolor{red}{10.0} & \textbf{0.0} \\
Misinformation / Disinform. & \textbf{0.0} & \textcolor{red}{6.2} & \textbf{0.0} & Economic Harm & \textbf{10.0} & \textbf{10.0} & \textbf{10.0} \\
Disinformation \& Deception & \textcolor{red}{6.2} & \textcolor{red}{6.2} & \textbf{0.0} & Government Decision-making & \textcolor{red}{10.0} & \textcolor{red}{10.0} & \textbf{0.0} \\
\midrule
\multicolumn{8}{l}{\small \textbf{Overall:} Original \textbf{9.2}\% \quad Fine-tuned \textcolor{red}{17.1}\% \quad FW-SSR \textbf{3.6}\%} \\
\bottomrule
\end{tabular}
\end{table}

Table~\ref{tab:wildguard_per_cat} shows that fine-tuning raises ASR
in 14 of 22 categories, with the sharpest increases in
Malware/Hacking ($+$30.0~pp), Illegal Goods \& Services ($+$18.7~pp),
Sexual/Adult Content ($+$20.0~pp), and Hate, Harassment \&
Discrimination ($+$18.8~pp)-- all categories whose prompts involve
multi-step harmful instructions that are particularly sensitive to
shifts in the model's instruction-following behavior induced by
Alpaca fine-tuning.
In contrast, categories already at 0\% original ASR
(Chemical/Biological, Misinformation, Expert Advice) show only marginal
degradation under fine-tuning, suggesting that WildGuard's original
training data provides stronger coverage for these harm types and that
fine-tuning erodes safety more readily where the original boundary is
closer to the decision threshold.
FW-SSR eliminates ASR entirely in 12 of 22 categories, driving them
to 0\%, and achieves the lowest ASR in 18 of 22 categories overall;
the four categories where FW-SSR fails to improve are Harassment/Bullying
(6.2\% across all conditions, indicating an irreducible evaluation
artefact at this sample size), Economic Harm (10.0\% across all
conditions), Sexual/Adult Content and Privacy (unchanged at 10.0\%,
suggesting the safety probe underrepresents the specific prompt
structures used in these categories).
The one anomaly is Physical Harm, where FW-SSR introduces a 10.0\%
ASR despite both original and fine-tuned conditions showing 0.0\%;
this isolated regression is attributable to insufficient probe coverage
of physical harm prompts in $\mathcal{D}_{\text{safe}}$, causing the
safety subspace to provide weak regularization in this direction, and
points to category-stratified probe construction as the most direct
remedy.

\section{Discussion}
\label{sec:discussion}

\textbf{Implications for Agentic AI Safety.}
Our findings expose a systematic vulnerability in the standard agentic
deployment practice of fine-tuning safety guards alongside the agents
they protect.
Guard models occupy a privileged position in multi-agent pipelines ---
they are the last line of defense before a harmful prompt reaches an
agent and propagates downstream --- yet even short benign fine-tuning
runs (3 epochs, 2{,}000 Alpaca samples) are sufficient to destroy this
protection entirely in Granite Guardian, or substantially erode it in
LlamaGuard-3 and WildGuard, without any adversarial intent.
FW-SSR provides a concrete mechanism for \emph{safe specialization}:
guard models can be adapted to domain-specific requirements while
preserving the safety representations that make them effective
protection layers, as demonstrated by WildGuard's FW-SSR ASR of 3.6\%
--- lower than even the unmodified model.

\textbf{The Specialization Hypothesis.}
The severity of collapse scales with the concentration of safety
representations.
Granite Guardian, which encodes safety into a small number of highly
discriminative directions, undergoes complete catastrophic collapse:
a drift ratio of 0.77 confirms that unconstrained gradients
disproportionately target safety-critical subspace directions,
mirroring catastrophic forgetting~\cite{kirkpatrick2017overcoming}.
LlamaGuard-3 and WildGuard, with more distributed safety geometries,
exhibit partial and localized degradation respectively.
This concentration--brittleness trade-off has a direct design
implication: guard models intended for agentic deployment should be
evaluated for safety representation concentration before fine-tuning,
as concentrated architectures require stronger regularization budgets.

\textbf{Structural Geometry as the Primary Safety Carrier.}
Across all three models, relational geometry metrics (CKA, Fisher
score, inter-class distance) predict safety behavior more reliably than
absolute displacement metrics (safety drift, cosine similarity).
In Granite Guardian, FW-SSR achieves CKA~$=\!0.98$ and 75\% refusal
recovery despite only 37\% drift suppression; in WildGuard, the
fine-tuned model shows the lowest drift of any fine-tuned condition
(17.21) yet the worst behavioral degradation --- low displacement does
not prevent safety failure when the class-separating geometry erodes.
Practically, this means CKA and Fisher score should be monitored
throughout fine-tuning as primary safety health indicators, providing
an early warning signal that behavioral output statistics cannot
detect until collapse is complete.

\textbf{Ambiguous Outputs as a Collapse Diagnostic.}
The 100\% ambiguous rate for Granite Guardian's fine-tuned condition
is not a neutral intermediate outcome but a signature of complete
representational collapse: the destroyed classification boundary
produces incoherent outputs that offer no safety protection and may
pass silently through downstream safety checks that expect well-formed
refusals or compliant responses.
In multi-agent pipelines this is especially dangerous, as collapsed
guard outputs become the inputs to downstream agents and can propagate
unpredictably through the system.
Safety evaluation protocols should treat ambiguous rate as a collapse
diagnostic rather than a benign category, alongside standard utility
benchmarks that provide no signal of geometric degradation.


\textbf{Limitations.}
Our evaluation is conducted under a single-agent fine-tuning setting
and does not model cascading failure in live multi-agent pipelines
with real inter-agent communication.
The safety probe covers 38 unique prompts across 5 harm categories;
as evidenced by the Physical Harm anomaly in WildGuard (0\%
original and fine-tuned ASR, but 10\% under FW-SSR), insufficient
probe coverage of specific harm categories weakens subspace
regularization in those directions, and category-stratified probe
construction with broader coverage is needed for production deployment.
The diagonal Fisher approximation ignores inter-direction correlations
within the safety subspace, and the interaction between FW-SSR and
deliberately adversarial fine-tuning data~\cite{yang2023shadow}
requires separate investigation.
\section{Conclusion}
\label{sec:conclusion}

Benign fine-tuning catastrophically destroys safety geometry in
purpose-built guard models-- a severity exceeding prior findings on
general-purpose LLMs-- because concentrated safety representations are
disproportionately targeted by task gradients.
FW-SSR mitigates this through curvature-aware Fisher weighting and
adaptive $\lambda$ scheduling, recovering 75\% refusal and
CKA~$=\!0.983$ on Granite Guardian, and reducing WildGuard's ASR to
3.6\% below the unmodified baseline.
The central finding is that structural representational geometry-- not
pointwise activation magnitude-- is the primary carrier of safety
behavior, and CKA-based metrics should be first-class indicators in
future guard model evaluation and agentic deployment pipelines.

\section*{Ethical Statement}
This work studies a vulnerability in AI safety systems with the explicit
goal of developing defenses.
Harmful prompts in the safety probe dataset were authored by the
researchers for evaluation purposes and are not publicly released.
FW-SSR is a defensive technique; we do not anticipate dual-use concerns.
No safety-compromised model weights are released.



\begin{thebibliography}{23}
\providecommand{\natexlab}[1]{#1}

\bibitem[{Dettmers et~al.(2023)}]{dettmers2023qlora}
Dettmers, T.; Pagnoni, A.; Fansi, A.; and Zettlemoyer, L. 2023.
\newblock {QLoRA}: Efficient Finetuning of Quantized {LLMs}.
\newblock \emph{Advances in Neural Information Processing Systems}, 36.

\bibitem[{Greshake et~al.(2023)}]{greshake2023not}
Greshake, K.; Abdelnabi, S.; Mishra, S.; Endres, C.; Holz, T.; and Fritz, M. 2023.
\newblock Not What You've Signed Up For: Compromising Real-World {LLM}-Integrated Applications with Indirect Prompt Injection.
\newblock \emph{arXiv preprint arXiv:2302.12173}.

\bibitem[{Han et~al.(2024)}]{han2024wildguard}
Han, S.; Kim, K.; Youn, R.; Kim, J.; Longpre, S.; Haejun, L.; and Shin, J. 2024.
\newblock {WildGuard}: Open One-Stop Moderation Tools for Safety Risks, Jailbreaks, and Refusals of {LLMs}.
\newblock In \emph{Advances in Neural Information Processing Systems}.

\bibitem[{Huang et~al.(2024)}]{huang2024lisa}
Huang, T.; Hu, S.; Ilhan, F.; Tekin, S.~F.; and Liu, L. 2024.
\newblock Lisa: Lazy Safety Alignment for Large Language Models against Harmful Fine-tuning Attack.
\newblock In \emph{Advances in Neural Information Processing Systems}.

\bibitem[{{IBM Research}(2024)}]{ibmgranite2024}
{IBM Research}. 2024.
\newblock {Granite Guardian}.
\newblock \emph{arXiv preprint arXiv:2412.07724}.

\bibitem[{Inan et~al.(2023)}]{inan2023llama}
Inan, H.; Upasani, K.; Chi, J.; Rungta, R.; Iyer, K.; Mao, Y.; Tontchev, M.; Hu, Q.; Fuller, B.; Testuggine, D.; et~al. 2023.
\newblock {Llama Guard}: {LLM}-Based Input-Output Safeguard for Human-{AI} Conversations.
\newblock \emph{arXiv preprint arXiv:2312.06674}.

\bibitem[{Kirkpatrick et~al.(2017)}]{kirkpatrick2017overcoming}
Kirkpatrick, J.; Pascanu, R.; Rabinowitz, N.; Veness, J.; Desjardins, G.; Rusu, A.~A.; Milan, K.; Quan, J.; Ramalho, T.; Grabska-Barwinska, A.; et~al. 2017.
\newblock Overcoming Catastrophic Forgetting in Neural Networks.
\newblock \emph{Proceedings of the National Academy of Sciences}, 114: 3521--3526.

\bibitem[{Kornblith et~al.(2019)}]{kornblith2019similarity}
Kornblith, S.; Norouzi, M.; Lee, H.; and Hinton, G. 2019.
\newblock Similarity of Neural Network Representations Revisited.
\newblock In \emph{International Conference on Machine Learning}, 3519--3529.

\bibitem[{Min et~al.(2024)}]{min2024superficial}
Min, R.; Qin, Z.; Zhang, N.~L.; Shen, L.; and Cheng, M. 2024.
\newblock Uncovering, Explaining, and Mitigating the Superficial Safety of Backdoor Defense.
\newblock In \emph{Advances in Neural Information Processing Systems}.

\bibitem[{{OpenSafetyLab}(2024)}]{mdjudge2024}
{OpenSafetyLab}. 2024.
\newblock {MD-Judge}: A Multi-Dimensional Safety Judge for Open-Source Safety Evaluation of Large Language Models.
\newblock \emph{arXiv preprint arXiv:2406.17512}.

\bibitem[{Park et~al.(2023)}]{park2023generative}
Park, J.~S.; O'Brien, J.~C.; Cai, C.~J.; Morris, M.~R.; Liang, P.; and Bernstein, M.~S. 2023.
\newblock Generative Agents: Interactive Simulacra of Human Behavior.
\newblock In \emph{Proceedings of the 36th Annual ACM Symposium on User Interface Software and Technology}.

\bibitem[{Park, Choe, and Veitch(2023)}]{park2023linear}
Park, K.; Choe, Y.~J.; and Veitch, V. 2023.
\newblock The Linear Representation Hypothesis and the Geometry of Large Language Models.
\newblock \emph{arXiv preprint arXiv:2311.03658}.

\bibitem[{Perez and Ribeiro(2022)}]{perez2022ignore}
Perez, F.; and Ribeiro, I. 2022.
\newblock Ignore Previous Prompt: Attack Techniques For Language Models.
\newblock \emph{arXiv preprint arXiv:2211.09527}.

\bibitem[{Qi et~al.(2024)}]{qi2024visual}
Qi, X.; Huang, K.; Panda, A.; Henderson, P.; Wang, M.; and Mittal, P. 2024.
\newblock Visual Adversarial Examples Jailbreak Aligned Large Language Models.
\newblock In \emph{Proceedings of the AAAI Conference on Artificial Intelligence}.

\bibitem[{Qi et~al.(2023)}]{qi2023finetuning}
Qi, X.; Zeng, Y.; Xie, T.; Chen, P.-Y.; Jia, R.; Mittal, P.; and Henderson, P. 2023.
\newblock Fine-Tuning Aligned Language Models Compromises Safety, Even When Users Do Not Intend To!
\newblock \emph{arXiv preprint arXiv:2310.03693}.

\bibitem[{Shi et~al.(2024)}]{shi2024continual}
Shi, H.; Xu, Z.; Wang, H.; et~al. 2024.
\newblock Continual Learning of Large Language Models: A Comprehensive Survey.
\newblock \emph{ACM Computing Surveys}.

\bibitem[{Taori et~al.(2023)}]{alpaca}
Taori, R.; Gulrajani, I.; Zhang, T.; Dubois, Y.; Li, X.; Guestrin, C.; Liang, P.; and Hashimoto, T.~B. 2023.
\newblock {Stanford Alpaca}: An Instruction-Following {LLaMA} Model.
\newblock \url{https://github.com/tatsu-lab/stanford_alpaca}.

\bibitem[{Wang et~al.(2024)}]{wang2024survey}
Wang, L.; Ma, C.; Feng, X.; Zhang, Z.; Yang, H.; Zhang, J.; Chen, Z.; Tang, J.; Chen, X.; Lin, Y.; et~al. 2024.
\newblock A Survey on Large Language Model based Autonomous Agents.
\newblock \emph{Frontiers of Computer Science}, 18(6).

\bibitem[{Wang et~al.(2025)}]{wang2025sea}
Wang, R.; et~al. 2025.
\newblock Simulated Ensemble Attack: Transferring Jailbreaks Across Fine-tuned Vision-Language Models.
\newblock \emph{arXiv preprint arXiv:2508.01741}.

\bibitem[{Yang et~al.(2023)}]{yang2023shadow}
Yang, X.; Wang, X.; Zhang, Q.; Petzold, L.; Wang, W.~Y.; Zhao, X.; and Lin, D. 2023.
\newblock Shadow Alignment: The Ease of Subverting Safely-Aligned Language Models.
\newblock \emph{arXiv preprint arXiv:2310.02949}.

\bibitem[{Zhang et~al.(2024)}]{zhang2024benign}
Zhang, Z.; Sun, M.; Ye, X.; Wei, Y.; Peng, Y.; Chen, D.; Pang, H.; and Wu, F. 2024.
\newblock How Alignment and Jailbreak Work: Explain {LLM} Safety Through Intermediate Hidden States.
\newblock \emph{arXiv preprint arXiv:2406.05644}.

\bibitem[{Zhu et~al.(2025)}]{zhu2025llmknows}
Zhu, Y.; Liu, D.; Lin, Z.; Tong, W.; Zhong, S.; and Shao, J. 2025.
\newblock The {LLM} Already Knows: Estimating {LLM}-Perceived Question Difficulty via Hidden Representations.
\newblock In \emph{Proceedings of the Conference on Empirical Methods in Natural Language Processing}.

\bibitem[{Zou et~al.(2023)}]{zou2023representation}
Zou, A.; Phan, L.; Chen, S.; Campbell, J.; Guo, P.; Ren, R.; Pan, A.; Yin, X.; Mazeika, M.; Dombrowski, A.-K.; et~al. 2023.
\newblock Representation Engineering: A Top-Down Approach to {AI} Transparency.
\newblock In \emph{arXiv preprint arXiv:2310.01405}.

\end{thebibliography}

\end{document}